\documentclass[runningheads]{llncs}

\usepackage{eccv}
\usepackage{eccvabbrv}
\usepackage{graphicx}
\usepackage{booktabs}
\usepackage{tabularx}
\usepackage{array}
\usepackage{amsmath,amssymb,mathtools}
\usepackage{xcolor}
\usepackage{xspace}
\usepackage{microtype}
\usepackage{float}
\usepackage{marvosym}
\usepackage[section]{placeins}
\usepackage[accsupp]{axessibility}
\usepackage[breaklinks,colorlinks,citecolor=eccvblue,linkcolor=eccvblue,urlcolor=eccvblue]{hyperref}

\makeatletter
\renewcommand{\@fnsymbol}[1]{\ensuremath{\ifcase#1\or\star\or\text{\Letter}\or
  {\star\star}\or \dagger\or \ddagger\or
  \mathchar "278\or \mathchar "27B\or \|\or **\or \dagger\dagger
  \or \ddagger\ddagger \else\@ctrerr\fi}}
\makeatother

\newcommand{\eos}{\mathrm{EOS}}
\newcommand{\prompt}[1]{\texttt{<#1>}}

\newcommand{\teaminfo}[3]{%
  \vspace{0.2mm}
  \begin{center}
  \renewcommand{\arraystretch}{0.92}%
  \setlength{\tabcolsep}{3pt}%
  \begin{tabularx}{\linewidth}{@{}>{\bfseries}lX@{}}
    \toprule
    Method & #1 \\
    Authors & #2 \\
    Affiliations & #3 \\
    \bottomrule
  \end{tabularx}
  \end{center}
  \vspace{0.2mm}
}

\title{Report of the 8th LSVOS Challenge:\\
Mobile Unified Multimodal Understanding}
\titlerunning{Report of the 8th LSVOS Challenge: MUMU Track}

\author{Chang Liu\thanks{Organizers of the MUMU track of the 8th LSVOS Challenge at ECCV 2026. The remaining authors are members of the top three teams. Chang Liu is with Shanghai University of Finance and Economics, and Tong Yang is with Transsion.}, \quad
Henghui Ding\textsuperscript{$\star$}\thanks{Corresponding author: Henghui Ding (\email{henghui.ding@gmail.com}), Institute of Big Data, Fudan University, Shanghai, China.}, \quad
Tong Yang\textsuperscript{$\star$}\\
\vspace{0.3cm}
Dayoung~Kil, \quad Seong-heum~Kim, \quad
Qilin~Xiang, \quad Ronghao~Zheng, \quad Ziyang~Zhang, \quad Shanlin~Jiang, \quad Ruohan~Shi, \quad Qilin~Fan, \quad
Chengfeng~Qiu, \quad Kaifeng~Wei}
\authorrunning{Liu et al.}
\institute{}

\begin{document}
\maketitle

\begingroup
\centering
{\url{https://lsvos.github.io/mumu}\par}
\endgroup

\begin{abstract}
This report summarizes the Mobile Unified Multimodal Understanding (MUMU) track of the 8th Large-scale Video Object Segmentation (LSVOS) Challenge, held in conjunction with ECCV 2026. MUMU asks for a single unified model that jointly performs multi-concept image tagging over quality, scene, and event vocabularies, open-vocabulary object detection, and concise image captioning, within a budget of 0.5B parameters and 8\,GB of peak inference memory. Submissions are ranked by a weighted task score scaled by an efficiency factor. We describe the tasks, rules, and evaluation protocol, report the final results, and review the methods of the top three teams. All three leading solutions build on the compact Florence-2-base model, but extend it in different ways: lightweight tagging heads with cross-task inference refinement, a shared low-rank adapter trained on coverage-aware multi-source data with task-dependent decoding, and a training-free pipeline that maps captions to tags and applies detection test-time augmentation. The results show that a compact prompt-based vision--language model can serve all three outputs when supervision, decoding, and cross-task cues are designed for each task, while open-vocabulary detection on novel categories remains the main bottleneck.
\end{abstract}

\section{Introduction}
\label{sec:mumu-intro}

Visual understanding on mobile and other resource-constrained devices rarely reduces to a single task. A practical system may need to describe the global content and quality of an image, localize the objects it contains, and summarize the scene in natural language. Maintaining a dedicated model for each capability multiplies parameters, memory, and maintenance cost, and prevents the models from using each other's predictions. Unified vision--language models address this by casting heterogeneous tasks as conditional sequence generation over a shared representation~\cite{chen2022unified,wang2022ofa,lu2023unifiedio,xiao2024florence2}. Many recent multimodal models, however, couple image encoders with billion-parameter language models~\cite{li2023blip2}, far beyond what can be deployed on a mobile budget.

The Mobile Unified Multimodal Understanding (MUMU) track studies this problem directly~\cite{mumu2026}. Participants must design a \emph{single} unified model that simultaneously performs multi-concept image tagging, open-vocabulary object detection, and image captioning under a constrained computational budget. Independent task-specific models are not permitted; adapters and low-rank modules are allowed only inside the unified architecture. The challenge emphasizes algorithmic efficiency, architectural unification, and deployability, and measures complexity with PC-side compute statistics instead of on-device execution.

The three outputs place different demands on a shared model. Tagging predicts closed-vocabulary label sets in which rare quality, scene, and event concepts all count equally under macro-averaging. Detection predicts a variable-size set of labeled boxes and must generalize to categories hidden from participants. Captioning must produce a fluent, image-grounded English sentence under a strict length limit. A compact model must therefore balance supervision, decoding, and post-processing across tasks whose errors have different costs, and no off-the-shelf compact model natively covers all three.

MUMU is held as an additional track of the 8th LSVOS Challenge in conjunction with ECCV 2026 in Malm\"o, Sweden~\cite{lsvos2026}. Alongside it, the three main tracks of the challenge address complex video object segmentation on MOSEv2 and text- and audio-guided referring video object segmentation on MeViSv2~\cite{ding2023mose,ding2025mosev2,ding2023mevis,ding2025mevisv2,GRES,ding2026multimodal,ding2026grex}. While these tracks focus on dense spatio-temporal grounding, MUMU complements them with efficient image-level multimodal understanding.

In this report, we describe the MUMU tasks, rules, and evaluation protocol, present the official results, and summarize the solutions of the top three teams from their technical reports. We then compare their recurring design choices, including the shared backbone, how tagging is added to a prompt-based model, cross-task reuse of predictions, and task-specific decoding, and discuss open directions.

\section{The MUMU Track}
\label{sec:challenge-overview}

\subsection{Tasks}
\label{sec:mumu-tasks}

For every test image, a submission must return the outputs of three tasks, all produced by the same unified model through a single inference call.

\noindent\textbf{Task A: Multi-Concept Image Tagging.}
The model predicts semantic concept labels in three namespaces: image quality attributes (\eg, blur, noise, exposure, and compression artifacts), scene categories (\eg, indoor, office, kitchen, street, and stadium), and event categories (\eg, concert, wedding, sports event, protest, and festival). Labels must be drawn from the published closed vocabulary, which contains 10 quality, 35 scene, and 18 event labels.
Each namespace is a multi-label prediction, and empty lists are allowed.

\noindent\textbf{Task B: Open-Vocabulary Object Detection.}
The model detects objects and returns, for each detection, a free-form lowercase text label, a confidence score in $[0,1]$, and a bounding box in absolute pixel coordinates of the original image. At most 300 detections are accepted per image. Novel detection categories are hidden in the final evaluation, so the model must generalize beyond the categories seen during training.

\noindent\textbf{Task C: Efficient Image Captioning.}
The model generates a concise, single-line English description of each image. A caption must be non-empty and contain at most 300 characters and at most 30 evaluator tokens, where tokens are counted after Unicode normalization and lowercasing.

\subsection{Datasets}
\label{sec:mumu-data}

Participants are encouraged to train on the public datasets listed in Table~\ref{tab:mumu-datasets}, which together cover the three tasks. Because most sources annotate only one namespace or task, constructing a training mixture with adequate label coverage is itself part of the challenge. Final rankings are computed on a held-out test set of 1{,}038 images whose annotations are not released.

\begin{table}[H]
  \caption{Recommended training datasets of the MUMU track.}
  \label{tab:mumu-datasets}
  \centering
  \scriptsize
  \setlength{\tabcolsep}{5pt}%
  \renewcommand{\arraystretch}{0.95}%
  \begin{tabularx}{\linewidth}{@{}l>{\raggedright\arraybackslash}Xl@{}}
    \toprule
    Output & Datasets & Purpose \\
    \midrule
    Quality    & KonIQ-10k~\cite{hosu2020koniq}, SPAQ~\cite{fang2020spaq}, LIVE Challenge~\cite{ghadiyaram2016livechallenge} & Image quality attributes \\
    Scene      & Places365~\cite{zhou2018places} & Scene recognition \\
    Event      & EventNet~\cite{ye2015eventnet}, NUS-WIDE~\cite{chua2009nuswide} & Event semantics \\
    Detection  & Objects365~\cite{shao2019objects365}, LVIS~\cite{gupta2019lvis}, Visual Genome~\cite{krishna2017visualgenome} & Open-vocabulary detection \\
    Captioning & COCO Captions~\cite{chen2015cococaptions}, Flickr30k~\cite{young2014flickr30k}, Conceptual Captions~\cite{sharma2018conceptual} & Caption generation \\
    \bottomrule
  \end{tabularx}
\end{table}

\subsection{Rules and Resource Constraints}
\label{sec:mumu-rules}

Each submission must use a single unified model. Adapters and LoRA modules are allowed within this architecture, but independent task-specific models are not, and all task heads must be integrated into the unified model. The model may contain at most 0.5B parameters and use at most 8\,GB of peak inference memory, measured on the PC side. Submissions declare their parameter count, GFLOPs for one forward pass at $224\times224$ resolution, and peak inference memory; declared values above the limits are rejected automatically. Each team also submits model weights and an inference script that returns tags, detections, and a caption from a single call. Top-ranked entries are reproduced by the organizers and disqualified if their results cannot be reproduced, if they use independent task-specific models, or if they exceed any resource constraint. The organizers provide a baseline built on the pretrained Florence-2-base model~\cite{xiao2024florence2} and Microsoft's public Florence-2 fine-tuning toolkit.

\subsection{Evaluation Protocol}
\label{sec:mumu-eval}

Task~A is evaluated by macro-F1 over the tag vocabulary, Task~B by novel-aware mAP over IoU thresholds from 0.50 to 0.95, and Task~C by a combination of CIDEr-D~\cite{vedantam2015cider}, SPICE~\cite{anderson2016spice}, and CLIPScore~\cite{hessel2021clipscore}:
\begin{equation}
  C = 0.4\,\mathrm{CIDEr} + 0.3\,\mathrm{SPICE} + 0.3\,\mathrm{CLIPScore}.
  \label{eq:mumu-caption}
\end{equation}
The three task scores, denoted $A$, $B$, and $C$, are combined into a weighted performance term $S$, which is scaled by an efficiency factor $E$ determined by model complexity:
\begin{equation}
  S = 0.3A + 0.4B + 0.3C, \qquad \mathrm{Final} = S\times E.
  \label{eq:mumu-final}
\end{equation}
Ties are broken by lower peak inference memory, then by lower parameter count, and finally by higher detection performance.

\subsection{Challenge Results}
\label{sec:mumu-results}

Table~\ref{tab:mumu-results} reports the final leaderboard. SSUfficient wins the track with a Final score of 17.34, followed by Team Spicy Gradient Descent with 16.72 and DDUN with 16.48. All three entries improve over the Florence-2-base baseline. The winning entry obtains the best detection and captioning scores, whereas Team Spicy Gradient Descent achieves the best tagging score, more than 11 points above the other two entries. Team Spicy Gradient Descent fine-tunes the shared model for all three tasks and obtains a lower caption score than the baseline. The first- and third-place entries keep the pretrained captioning pathway frozen and obtain the two highest caption scores. Detection scores remain low for every team, between 4.36 and 8.22, reflecting the difficulty of generalizing to hidden novel categories.

Table~\ref{tab:mumu-resources} summarizes model complexity. All three teams build on Florence-2-base, so their models contain between 231M and 240M parameters, less than half of the 0.5B limit, and their efficiency factors are close. Peak inference memory ranges from about 0.6\,GB to 7.5\,GB, below the 8\,GB limit in all cases.

\begin{table}[H]
  \caption{MUMU final test-set results. $A$: tagging macro-F1; $B$: novel-aware detection mAP@[0.5:0.95]; $C$: caption score (Eq.~\ref{eq:mumu-caption}); $E$: efficiency factor; $\mathrm{Final}=(0.3A+0.4B+0.3C)\times E$. Bold marks the best score among ranked entries. $^\dagger$Florence-2-base evaluated with the official baseline inference script.}
  \label{tab:mumu-results}
  \centering
  \scriptsize
  \setlength{\tabcolsep}{7pt}%
  \renewcommand{\arraystretch}{0.95}%
  \begin{tabular}{@{}clrrrrr@{}}
    \toprule
    Rank & Team & $A$ & $B$ & $C$ & $E$ & Final \\
    \midrule
    1 & SSUfficient                 & 27.18 & \textbf{8.22} & \textbf{47.34} & 0.676 & \textbf{17.3409} \\
    2 & Team Spicy Gradient Descent & \textbf{38.63} & 4.36 & 34.89 & 0.702 & 16.7172 \\
    3 & DDUN                        & 26.49 & 5.45 & 44.05 & \textbf{0.706} & 16.4815 \\
    \midrule
    -- & Baseline (Florence-2-base)$^\dagger$ & 21.15 & 4.77 & 44.05 & 0.706 & 15.1568 \\
    \bottomrule
  \end{tabular}
\end{table}

\begin{table}[H]
  \caption{Model complexity of the top three entries. All three build on Florence-2-base; Adaptation lists what each entry adds to it. Mem.: peak inference memory.}
  \label{tab:mumu-resources}
  \centering
  \scriptsize
  \setlength{\tabcolsep}{5pt}%
  \renewcommand{\arraystretch}{0.95}%
  \begin{tabular}{@{}llrr@{}}
    \toprule
    Team & Adaptation & Params (M) & Mem. (GB) \\
    \midrule
    SSUfficient                 & Tagging heads             & 239.17 & 4.5 \\
    Team Spicy Gradient Descent & Fused LoRA                & 232.25 & 7.5 \\
    DDUN                        & None (training-free)      & 231.41 & 0.6 \\
    \bottomrule
  \end{tabular}
\end{table}

\section{Top Solutions in the MUMU Track}
\label{sec:mumu-solutions}

\subsection{1st Place: SSUfficient}
\label{sec:mumu1}

\teaminfo{EUMU: Efficient Unified Multimodal Understanding}{Dayoung Kil$^{1}$, Seong-heum Kim$^{1}$}{$^{1}$Department of Intelligent Semiconductors, Soongsil University, Seoul, Republic of Korea}

\begin{figure}[t]
  \centering
  \includegraphics[width=\linewidth]{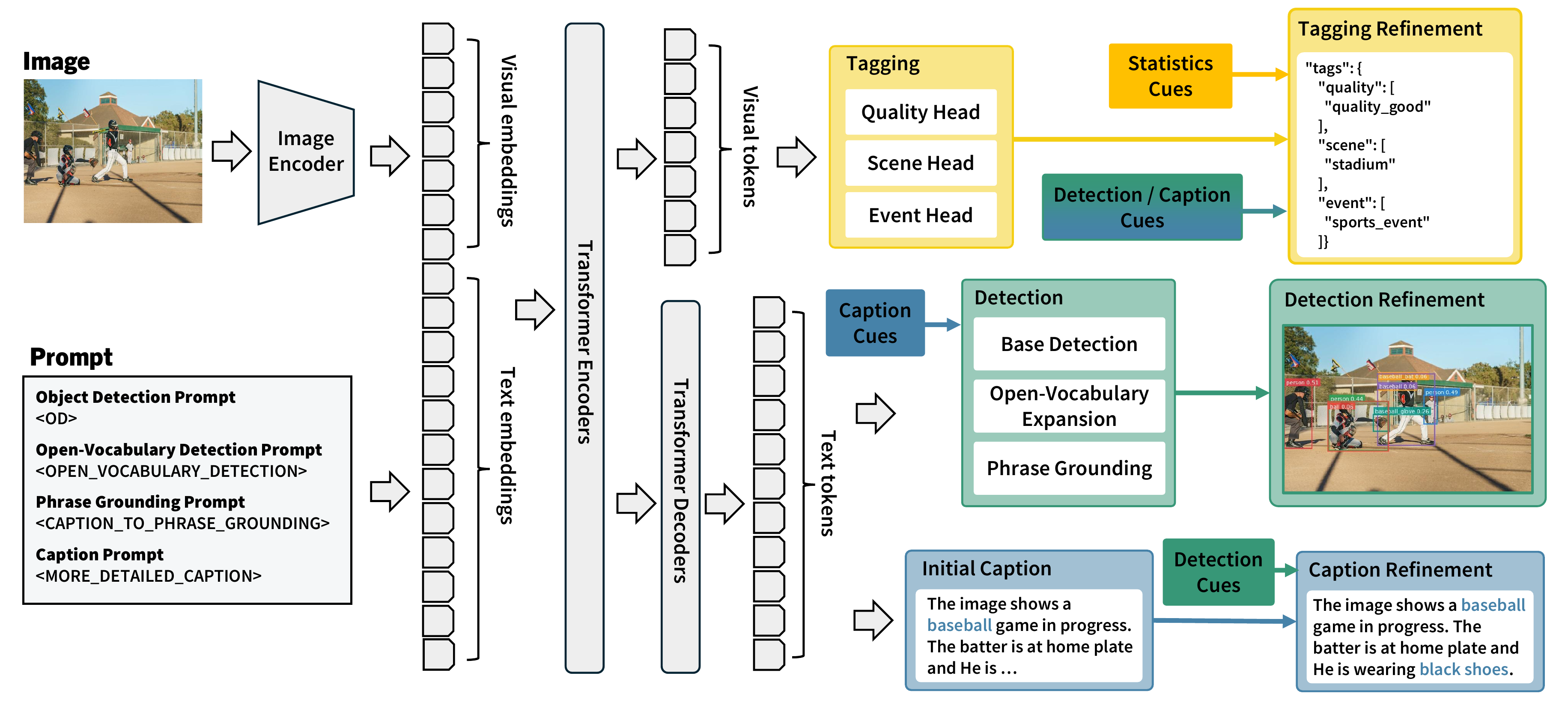}
  \caption{Overview of EUMU (SSUfficient). One Florence-2-base model is shared across tasks; lightweight heads predict tags from shared visual features, and task outputs are reused as cross-task cues for inference refinement.}
  \label{fig:mumu1-overview}
\end{figure}

EUMU shares a single frozen Florence-2-base model~\cite{xiao2024florence2} across tagging, detection, and captioning. As shown in Fig.~\ref{fig:mumu1-overview}, the input image is encoded once by the image encoder and transformer encoder into a shared visual representation. Detection and captioning use task prompts and the pretrained decoder without fine-tuning, while three lightweight tagging heads operate directly on the shared visual features. Instead of running the tasks independently, EUMU applies task-aware inference refinement in a directed sequence: it first obtains initial tags, base detections, and a context caption; the caption then expands the detection candidates; the final detections refine the caption; and lastly, image statistics, the final caption, and the final detections refine the tags.

\paragraph{Lightweight tagging heads.}
Florence-2 offers no native prompt for multi-concept tagging. EUMU therefore adds three MLP heads for quality, scene, and event prediction on top of the shared visual features. Each head produces label-wise scores from which the initial tags are obtained. Because tagging bypasses the transformer decoder, all three tag types are predicted without modifying the pretrained multimodal model. The heads are trained on the official MUMU training data, supplemented with samples constructed from the same split to address sparse quality and event labels: synthetic degradations such as blur, noise, low light, and overexposure provide additional quality examples, and event-related cues extracted from training captions provide additional event labels.

\paragraph{Caption-guided open-vocabulary detection.}
EUMU first runs base object detection with the \prompt{OD} prompt to obtain boxes and labels. To recover objects missed at this stage, it extracts object and phrase cues from the context caption. These cues drive two further passes of the same model: open-vocabulary detection (\prompt{OPEN\_\allowbreak VOCABULARY\_\allowbreak DETECTION}) detects additional objects named in the caption, and phrase grounding (\prompt{CAPTION\_\allowbreak TO\_\allowbreak PHRASE\_\allowbreak GROUNDING}) localizes caption phrases. Candidates from the three stages are merged, labels are normalized to the evaluation vocabulary, invalid boxes are removed, and non-maximum suppression (NMS) suppresses duplicates. The remaining boxes, labels, and confidence scores form the final detections.

\paragraph{Detection-refined captioning.}
The context caption is generated with the \prompt{MORE\_\allowbreak DETAILED\_\allowbreak CAPTION} prompt. After the final detections are available, EUMU compares the detected object labels with the words in the caption and uses their overlap as an additional cue to refine it, so that the caption better reflects the detected objects. The refined caption is then post-processed to satisfy the length limits while preserving sentence completeness: overly long outputs are truncated, incomplete sentence endings caused by truncation are removed, redundant whitespace and line breaks are deleted, and sentence-ending punctuation is added when needed.

\paragraph{Cue-based tag refinement.}
Each label group is refined with the cues most relevant to it. Quality labels are tied to low-level image characteristics, so the head scores are complemented by image statistics such as brightness distribution, contrast, gradient, noise, and compression artifacts; for example, brightness statistics refine low-light, underexposure, and overexposure predictions. Scene predictions are refined with location cues from the final caption together with the detected objects, which allows a generic prediction such as \emph{outdoor} to be replaced by a more specific scene category. Event predictions combine action cues from the caption with object cues from the detections, such as a stage and a microphone for a performance or a race number and a finish line for a sports event. Event predictions inconsistent with the caption, scene tags, and detections are removed.

\subsection{2nd Place: Team Spicy Gradient Descent}
\label{sec:mumu2}

\teaminfo{Coverage-Aware Supervision and Task-Dependent Decoding}{Qilin Xiang$^{1}$, Ronghao Zheng$^{1}$, Ziyang Zhang$^{1}$, Shanlin Jiang$^{1}$, Ruohan Shi$^{2}$, Qilin Fan$^{1}$}{$^{1}$School of Big Data and Software Engineering, Chongqing University, Chongqing, China; $^{2}$Harbin Institute of Technology, Weihai, China}

\begin{figure}[!htb]
  \centering
  \includegraphics[width=0.92\linewidth]{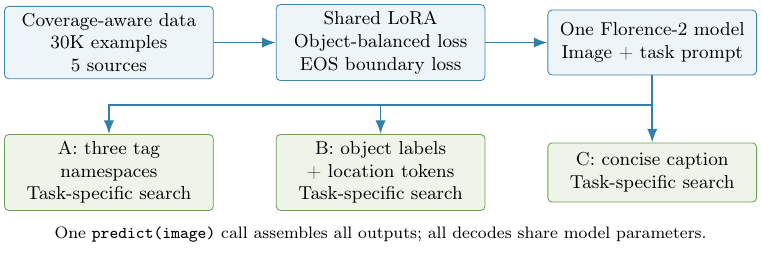}
  \caption{Overview of the Team Spicy Gradient Descent system. Coverage-aware sampling and namespace-specific supervision build the training mixture; one Florence-2 model with a shared LoRA is decoded with task-dependent policies.}
  \label{fig:mumu2-overview}
\end{figure}

The method adapts one Florence-2-base model~\cite{xiao2024florence2} with a single low-rank adapter shared by all three tasks, which are selected through task prompts. As shown in Fig.~\ref{fig:mumu2-overview}, the submitted model combines coverage-aware multi-source supervision with task-dependent decoding. For an image $x$, one call of the predictor returns
\begin{equation}
  F_\theta(x)=\bigl(\hat T_q,\hat T_s,\hat T_e,\hat D,\hat c\bigr),
  \label{eq:mumu2-predictor}
\end{equation}
namely the quality, scene, and event tag sets, a set of labeled boxes with confidences, and a caption.

\paragraph{Shared low-rank adaptation.}
A LoRA~\cite{hu2022lora} with rank 16 is applied to eligible linear layers. For a frozen weight $W_0$ and input $z$, it adds a trainable update $h=(W_0+\tfrac{\alpha}{r}UV)z$, and all tasks jointly update this adapter. Task~A uses a concise instruction requesting a compact JSON array for one namespace; Task~B uses the pretrained \prompt{OD} token and represents each object as its text label followed by four location tokens on a 1{,}000-bin grid; Task~C uses \prompt{CAPTION}. Each response is generated autoregressively as $p_\theta(y\mid x,p_t)=\prod_{j}p_\theta(y_j\mid x,p_t,y_{<j})$, so category names and coordinates are predicted in the same sequence. For export, non-tied LoRA matrices are fused into the dense model, while the small update to the output head, whose weights are tied to the input embeddings, is kept as a residual. The exported model has 232.25M parameters.

\paragraph{Coverage-aware multi-source supervision.}
Quality, scene, and event concepts come from different datasets, so label coverage depends on how sources are combined. The training mixture draws on five sources: KonIQ-10k~\cite{hosu2020koniq} images with deterministic distortion recipes for categorical quality labels, Places365~\cite{zhou2018places} annotations mapped to the 35 scene labels, NUS-WIDE~\cite{chua2009nuswide} user tags matched to the event vocabulary as weak labels, LVIS v1~\cite{gupta2019lvis} for detection, and Flickr30k~\cite{young2014flickr30k} for captions. The mixture is expanded from 10K to 30K examples, comprising 9{,}000 tagging, 10{,}500 detection, and 10{,}500 caption examples. Because a source image rarely annotates every field of Eq.~\eqref{eq:mumu2-predictor}, observed namespaces are represented by a supervision mask: an unannotated namespace is treated as unlabeled, not as a negative. Task sampling cycles each task to the size of the largest group, giving equal presentations of tagging, detection, and captioning in each effective epoch. On the official leaderboard, the 30K mixture raises the final score from 15.89 to 16.72, with the gain coming from tagging.

\paragraph{Task-dependent decoding.}
The same token distribution must be turned into label sets, ranked object sets, and sentences. For each prompted output $t$, the policy $\pi_t=(p_t,b_t,\ell_t,g_t)$ specifies the prompt, beam width, generation cap, and task-specific parsing and normalization:
\begin{equation}
  \hat z_t=g_t\!\left(\operatorname{Decode}_{\theta}(x,p_t;b_t,\ell_t)\right).
  \label{eq:mumu2-policy}
\end{equation}
Beam widths are selected by the metric of the processed task output instead of sequence likelihood. Because the local selection score is separable across tasks, each task's best width is chosen independently. The submitted model uses beams $(32,4,8)$ and caps $(64,256,64)$ for A/B/C. A local diagnostic motivates this separation: detection AP falls from 16.32 at beam 2 to 6.58 at beam 64, whereas tagging peaks at beam 64, only marginally above greedy decoding. Each decoded detection sequence receives a confidence $s(y)=\exp\bigl(\tfrac{1}{L}\sum_{j}\log p_\theta(y_j\mid x,p_B,y_{<j})\bigr)$ that its boxes inherit. Post-processing validates labels, discards degenerate boxes, removes same-label overlaps above IoU 0.82, keeps at most 20 boxes per image, and maps LVIS synonyms to canonical labels; captions are normalized and truncated to the evaluator limits.

\paragraph{Object-balanced sequence adaptation.}
The team further explores post-training stages on its local validation set. Token-level cross-entropy over-weights long class names and supervises the end-of-sequence (EOS) decision, which sets the number of output objects, with a single token. An object-normalized loss $\mathcal{L}_{\mathrm{obj}}$ therefore averages token losses within each object span and then across objects, and a boundary term
\begin{equation}
  \mathcal{L}_{\mathrm{boundary}}=-\frac{1}{N}\Bigl[\sum_{i=1}^{N-1}\log\bigl(1-p_{e_i}(\eos)\bigr)+\log p_{e_N}(\eos)\Bigr]
  \label{eq:mumu2-boundary}
\end{equation}
supervises continuation after non-final objects and termination after the last one; the two are combined as $\mathcal{L}_B=(\mathcal{L}_{\mathrm{obj}}+\lambda\mathcal{L}_{\mathrm{boundary}})/(1+\lambda)$ with $\lambda=4$. Pix2Seq-inspired sequence augmentation~\cite{chen2022pix2seq} adds synthetic noise objects during this stage. A subsequent stage applies mild appearance augmentation to tagging and captioning images and raises the local weighted score from 36.11 to 37.69.

\subsection{3rd Place: DDUN}
\label{sec:mumu3}

\teaminfo{CapMap-MS-TTA: Caption Keyword Mapping with Multi-Scale Flip Test-Time Augmentation}{Chengfeng Qiu$^{1}$, Kaifeng Wei$^{1}$}{$^{1}$Netease YiDun AI Lab, Hangzhou, China}

\begin{figure}[!htb]
  \centering
  \includegraphics[width=0.85\linewidth]{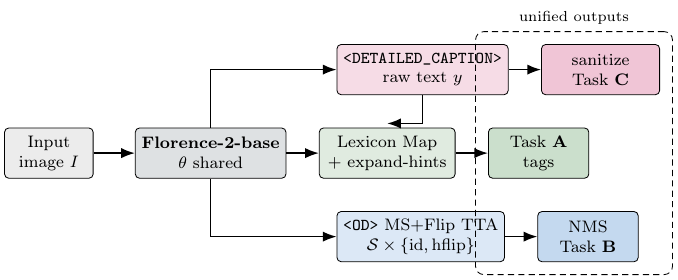}
  \caption{Overview of CapMap-MS-TTA (DDUN). One Florence-2-base checkpoint serves all tasks: Tasks~A and~C share the detailed-caption pathway, and Task~B uses multi-scale and flip detection TTA followed by NMS.}
  \label{fig:mumu3-overview}
\end{figure}

CapMap-MS-TTA is a training-free solution built on Florence-2-base~\cite{xiao2024florence2}, with 231.41M parameters and about 0.6\,GB of peak memory at batch size 1. All tasks share the same weights $\theta$ without fine-tuning. Florence-2 casts each task as conditional sequence generation, $o=\operatorname{Decode}(f_\theta(I,\tau))$, where $\tau$ is a task token. As shown in Fig.~\ref{fig:mumu3-overview}, Tasks~A and~C share a single detailed-caption pathway, while Task~B uses multi-view detection with test-time augmentation (TTA) followed by non-maximum suppression (NMS). The improvements come from prompt routing, deterministic post-processing, and detection TTA.

\paragraph{Task C: sanitized detailed captioning.}
For each image $I$, the \prompt{DETAILED\_\allowbreak CAPTION} token yields raw text $y$, and the submitted caption is
\begin{equation}
  c = \Pi_{\leq 30}\!\left(\mathrm{Trunc}_{\leq 300}\!\left(\mathrm{Norm}(y)\right)\right),
  \label{eq:mumu3-sanitize}
\end{equation}
where $\mathrm{Norm}$ collapses whitespace, $\mathrm{Trunc}_{\leq 300}$ enforces the character limit, and $\Pi_{\leq 30}$ greedily drops trailing tokens until at most 30 evaluator tokens remain. Decoding uses beam search with three beams and a fixed batch size of 1, since padding in larger batches changes the generations and breaks reproducibility. Detailed captions are often 50--80 words long, so truncation keeps the leading subject and setting at the cost of trailing attributes. Since captioning contributes the most to the final score, this pathway is kept unchanged.

\paragraph{Task A: caption keyword mapping with expand-hints.}
Tags are derived from the same detailed caption. A phrase-to-tag dictionary $\mathcal{M}$ maps phrases to labels of the closed vocabulary $\mathcal{V}=\mathcal{V}_q\cup\mathcal{V}_s\cup\mathcal{V}_e$, for example ``shopping mall'' to \texttt{shopping\_mall}. With $\tilde y$ the lower-cased caption,
\begin{equation}
  \mathrm{Map}(y;\mathcal{M})=\bigl\{\mathcal{M}(p)\;\big|\;p\in\mathrm{dom}(\mathcal{M}),\ p\text{ occurs in }\tilde y\text{ as a whole phrase}\bigr\}.
  \label{eq:mumu3-map}
\end{equation}
Whole-phrase matching avoids substring errors such as \texttt{ski} in ``skin'' or \texttt{mall} in ``small'', which otherwise produce many false scene tags. Quality tags additionally use image statistics: with $G$ the grayscale image and $\Delta$ a discrete Laplacian, the variance $v=\mathrm{Var}(\Delta G)$ and mean intensity $\mu=\mathrm{Mean}(G)$ yield heuristic tags $Q(v,\mu)\subseteq\mathcal{V}_q$ for blur, low light, and under- or overexposure. Starting from a base tag set $T_0$, the expand-hints stage only adds tags, $T=T_0\cup\mathrm{Map}(y;\mathcal{M})\cup Q(v,\mu)$, so it never reduces recall. A final cleanup removes \texttt{outdoor} when both \texttt{indoor} and \texttt{outdoor} are present, and otherwise adds \texttt{indoor} or \texttt{outdoor} when the scene tags contain an indoor- or outdoor-indicative category.

\paragraph{Task B: multi-scale and flip detection TTA.}
Florence-2 open detection (\prompt{OD}) is run on rescaled and horizontally flipped views. Let $\mathcal{S}=\{0.75,1.0,1.25\}$, let $\mathrm{H}$ denote horizontal flipping, and let $\Phi_s$ map boxes from the image $I_s$ at scale $s$ back to the original resolution. Flipped boxes are reflected as $\mathrm{H}^{-1}(b)=(W-x_2,y_1,W-x_1,y_2)$ before rescaling, and the proposal pool is
\begin{equation}
  \mathcal{P}=\bigcup_{s\in\mathcal{S}}\Bigl(\Phi_s\bigl(\mathrm{OD}(I_s)\bigr)\cup\Phi_s\bigl(\mathrm{H}^{-1}(\mathrm{OD}(\mathrm{H}(I_s)))\bigr)\Bigr).
  \label{eq:mumu3-pool}
\end{equation}
Smaller scales recover objects missed at native resolution, larger scales help in crowded scenes, and flipping adds complementary proposals for asymmetric objects. Label-aware NMS keeps a box unless a kept box with the same label overlaps it with IoU above 0.5, retaining at most 300 boxes, and malformed labels are discarded for schema validity. Because Florence-2 detection often returns no scores, boxes receive a constant confidence of 0.85, so the gain comes from geometric redundancy and NMS instead of score thresholding. Relative to the Florence-2-base baseline, expand-hints raises the tagging score from 21.15 to 26.49 and TTA raises the detection score from 4.77 to 5.45, while the caption score is unchanged.

\FloatBarrier

\section{Conclusion and Discussion}
\label{sec:mumu-conclusion}

All three leading solutions build on Florence-2-base, whose prompt-based sequence interface covers detection and captioning with one set of weights while leaving ample headroom under the 0.5B-parameter limit. They differ mainly in how they add multi-concept tagging, which Florence-2 does not support natively: SSUfficient trains lightweight heads on frozen shared features, Team Spicy Gradient Descent fine-tunes a shared LoRA to generate tag lists as text, and DDUN maps detailed captions to the closed vocabulary without training. Fine-tuning the shared model on coverage-aware tag supervision yields by far the highest tagging score but lowers caption quality, whereas the entries that keep the pretrained captioning pathway intact obtain the highest caption scores, which carry substantial weight in the final score.

A second recurring theme is that the tasks inform one another at inference time. Captions supply object names for open-vocabulary detection and phrase grounding, detections refine captions, and captions, detections, and image statistics refine quality, scene, and event tags. Teams also tune decoding and post-processing separately for each task, since beam widths, generation caps, and NMS thresholds that help one output can hurt another. Multi-scale and flip augmentation further increases detection recall without adding parameters.

Open-vocabulary detection on hidden novel categories remains the main bottleneck, with the best novel-aware mAP of only 8.22. The second- and third-place teams note that sequence-based detection in Florence-2 lacks calibrated per-box confidences and must decide through a single end-of-sequence token how many objects to emit, while the 30-token caption budget forces aggressive truncation of detailed captions. Future work could jointly train compact unified models across all three tasks under the resource budget, estimate region-level confidences within the shared decoder, and measure latency and memory on actual mobile hardware.

\bibliographystyle{splncs04}
\bibliography{main}

\begin{thebibliography}{10}
\providecommand{\url}[1]{\texttt{#1}}
\providecommand{\urlprefix}{URL }
\providecommand{\doi}[1]{https://doi.org/#1}

\bibitem{anderson2016spice}
Anderson, P., Fernando, B., Johnson, M., Gould, S.: {SPICE}: Semantic
  propositional image caption evaluation. In: ECCV. pp. 382--398 (2016)

\bibitem{chen2022pix2seq}
Chen, T., Saxena, S., Li, L., Fleet, D.J., Hinton, G.: {Pix2Seq}: A language
  modeling framework for object detection. In: ICLR (2022)

\bibitem{chen2022unified}
Chen, T., Saxena, S., Li, L., Lin, T.Y., Fleet, D.J., Hinton, G.: A unified
  sequence interface for vision tasks. In: NeurIPS (2022)

\bibitem{chen2015cococaptions}
Chen, X., Fang, H., Lin, T.Y., Vedantam, R., Gupta, S., Doll{\'a}r, P.,
  Zitnick, C.L.: Microsoft {COCO} captions: Data collection and evaluation
  server. arXiv preprint arXiv:1504.00325  (2015)

\bibitem{chua2009nuswide}
Chua, T.S., Tang, J., Hong, R., Li, H., Luo, Z., Zheng, Y.: {NUS-WIDE}: A
  real-world web image database from {National University of Singapore}. In:
  Proceedings of the ACM International Conference on Image and Video Retrieval
  (2009)

\bibitem{ding2026grex}
Ding, H., Liu, C., He, S., Jiang, X., Jiang, Y.G.: {GREx}: Generalized
  referring expression segmentation, comprehension, and generation.
  International Journal of Computer Vision  \textbf{134}(2), ~79 (2026)

\bibitem{ding2023mevis}
Ding, H., Liu, C., He, S., Jiang, X., Loy, C.C.: {MeViS}: A large-scale
  benchmark for video segmentation with motion expressions. In: ICCV (2023)

\bibitem{ding2023mose}
Ding, H., Liu, C., He, S., Jiang, X., Torr, P.H.S., Bai, S.: {MOSE}: A new
  dataset for video object segmentation in complex scenes. In: ICCV. pp.
  20224--20234 (2023)

\bibitem{ding2025mevisv2}
Ding, H., Liu, C., He, S., Ying, K., Jiang, X., Loy, C.C., Jiang, Y.G.:
  {MeViS}: A multi-modal dataset for referring motion expression video
  segmentation. IEEE Transactions on Pattern Analysis and Machine Intelligence
  \textbf{47}(12),  11400--11416 (2025)

\bibitem{ding2026multimodal}
Ding, H., Tang, S., He, S., Liu, C., Wu, Z., Jiang, Y.G.: Multimodal referring
  segmentation: A survey. International Journal of Computer Vision
  \textbf{134}(9), ~405 (2026)

\bibitem{ding2025mosev2}
Ding, H., Ying, K., Liu, C., He, S., Jiang, X., Jiang, Y.G., Torr, P.H.S., Bai,
  S.: {MOSEv2}: A more challenging dataset for video object segmentation in
  complex scenes. arXiv preprint arXiv:2508.05630  (2025)

\bibitem{fang2020spaq}
Fang, Y., Zhu, H., Zeng, Y., Ma, K., Wang, Z.: Perceptual quality assessment of
  smartphone photography. In: CVPR. pp. 3677--3686 (2020)

\bibitem{ghadiyaram2016livechallenge}
Ghadiyaram, D., Bovik, A.C.: Massive online crowdsourced study of subjective
  and objective picture quality. IEEE Transactions on Image Processing
  \textbf{25}(1),  372--387 (2016)

\bibitem{gupta2019lvis}
Gupta, A., Doll{\'a}r, P., Girshick, R.: {LVIS}: A dataset for large vocabulary
  instance segmentation. In: CVPR. pp. 5356--5364 (2019)

\bibitem{hessel2021clipscore}
Hessel, J., Holtzman, A., Forbes, M., {Le Bras}, R., Choi, Y.: {CLIPScore}: A
  reference-free evaluation metric for image captioning. In: EMNLP. pp.
  7514--7528 (2021)

\bibitem{hosu2020koniq}
Hosu, V., Lin, H., Sziranyi, T., Saupe, D.: {KonIQ-10k}: An ecologically valid
  database for deep learning of blind image quality assessment. IEEE
  Transactions on Image Processing  \textbf{29},  4041--4056 (2020)

\bibitem{hu2022lora}
Hu, E.J., Shen, Y., Wallis, P., Allen-Zhu, Z., Li, Y., Wang, S., Wang, L.,
  Chen, W.: {LoRA}: Low-rank adaptation of large language models. In: ICLR
  (2022)

\bibitem{krishna2017visualgenome}
Krishna, R., Zhu, Y., Groth, O., Johnson, J., Hata, K., Kravitz, J., Chen, S.,
  Kalantidis, Y., Li, L.J., Shamma, D.A., Bernstein, M.S., Fei-Fei, L.: Visual
  {Genome}: Connecting language and vision using crowdsourced dense image
  annotations. International Journal of Computer Vision  \textbf{123}(1),
  32--73 (2017)

\bibitem{lsvos2026}
{Large-scale Video Object Segmentation Workshop}: The 8th {LSVOS} challenge:
  Tracks and submission. \url{https://lsvos.github.io/} (2026), accessed August
  14, 2026

\bibitem{mumu2026}
{Large-scale Video Object Segmentation Workshop}: {MUMU Challenge 2026}: Mobile
  unified multimodal understanding. \url{https://lsvos.github.io/mumu} (2026)

\bibitem{li2023blip2}
Li, J., Li, D., Savarese, S., Hoi, S.: {BLIP-2}: Bootstrapping language-image
  pre-training with frozen image encoders and large language models. In: ICML
  (2023)

\bibitem{GRES}
Liu, C., Ding, H., Jiang, X.: {GRES}: Generalized referring expression
  segmentation. In: CVPR. pp. 23592--23601 (2023)

\bibitem{lu2023unifiedio}
Lu, J., Clark, C., Zellers, R., Mottaghi, R., Kembhavi, A.: {Unified-IO}: A
  unified model for vision, language, and multi-modal tasks. In: ICLR (2023)

\bibitem{shao2019objects365}
Shao, S., Li, Z., Zhang, T., Peng, C., Yu, G., Zhang, X., Li, J., Sun, J.:
  {Objects365}: A large-scale, high-quality dataset for object detection. In:
  ICCV. pp. 8430--8439 (2019)

\bibitem{sharma2018conceptual}
Sharma, P., Ding, N., Goodman, S., Soricut, R.: Conceptual captions: A cleaned,
  hypernymed, image alt-text dataset for automatic image captioning. In: ACL.
  pp. 2556--2565 (2018)

\bibitem{vedantam2015cider}
Vedantam, R., Zitnick, C.L., Parikh, D.: {CIDEr}: Consensus-based image
  description evaluation. In: CVPR. pp. 4566--4575 (2015)

\bibitem{wang2022ofa}
Wang, P., Yang, A., Men, R., Lin, J., Bai, S., Li, Z., Ma, J., Zhou, C., Zhou,
  J., Yang, H.: {OFA}: Unifying architectures, tasks, and modalities through a
  simple sequence-to-sequence learning framework. In: ICML (2022)

\bibitem{xiao2024florence2}
Xiao, B., Wu, H., Xu, W., Dai, X., Hu, H., Lu, Y., Zeng, M., Liu, C., Yuan, L.:
  {Florence-2}: Advancing a unified representation for a variety of vision
  tasks. In: CVPR. pp. 4818--4829 (2024)

\bibitem{ye2015eventnet}
Ye, G., Li, Y., Xu, H., Liu, D., Chang, S.F.: {EventNet}: A large scale
  structured concept library for complex event detection in video. In: ACM MM.
  pp. 471--480 (2015)

\bibitem{young2014flickr30k}
Young, P., Lai, A., Hodosh, M., Hockenmaier, J.: From image descriptions to
  visual denotations: New similarity metrics for semantic inference over event
  descriptions. Transactions of the Association for Computational Linguistics
  \textbf{2},  67--78 (2014)

\bibitem{zhou2018places}
Zhou, B., Lapedriza, A., Khosla, A., Oliva, A., Torralba, A.: Places: A 10
  million image database for scene recognition. IEEE Transactions on Pattern
  Analysis and Machine Intelligence  \textbf{40}(6),  1452--1464 (2018)

\end{thebibliography}

\end{document}